\documentclass[runningheads]{llncs}

\usepackage{accv}

\usepackage{accvabbrv}

\usepackage{graphicx}

\usepackage{booktabs}
\usepackage{array}
\usepackage{tabularx}
\usepackage{xcolor}
\usepackage{tcolorbox}
\usepackage{caption}
\usepackage{subcaption}
\usepackage{makecell}
\usepackage{float}
\usepackage[accsupp]{axessibility}  

\usepackage{hyperref}

\usepackage{orcidlink}

\begin{document}

\title{What Constitutes Food Similarity? Visual Representations and Retrieval in Indian Cuisine}

\titlerunning{What Constitutes Food Similarity?}

\author{Saransh Kumar Gupta\inst{1}\orcidlink{0009-0000-5887-2301}\thanks{Corresponding Author.} \and
Akshadha Sharma\inst{1} \and
Lipika Dey\inst{1}\orcidlink{0000-0003-3831-5545} \and
Partha Pratim Das\inst{1}\orcidlink{0000-0003-1435-6051} \and
Ramesh Jain\inst{2}\orcidlink{0000-0003-2373-4966}}

\authorrunning{S.K. Gupta et al.}


\institute{Ashoka University, India\\
\email{\{saransh.gupta,lipika.dey,partha.das\}@ashoka.edu.in}\\
\email{akshadha.sharma\_ug2023@ashoka.edu.in}\\
\url{https://www.ashoka.edu.in/} \and
Institute for Future Health, UC Irvine, USA\\
\email{jain\@ics.uci.edu}\\
\url{https://futurehealth.uci.edu/}}

\maketitle

\begin{abstract}
    Food similarity underlies retrieval, recommendation, substitution, and other food computing tasks, yet food representations are typically evaluated by identity preservation. We ask what happens to food similarity beyond identity: how different visual representations organize foods into relational neighborhoods, what culinary relationships these neighborhoods reveal or obscure, and how consistently these structures emerge across representations. Indian cuisine provides a challenging setting in which visually similar foods may have different culinary identities, while foods within a culinary family can vary substantially in appearance. Using an expanded Khana dataset and seven pretrained visual representations spanning convolutional, transformer-based, self-supervised, and image-text aligned models, we evaluate image-to-image retrieval and image-to-dish neighborhoods, examining culinary-family structure, contrast cases, and cross-representation agreement. We also aggregate image representations to compare dish-level visual similarity across models and against recipe-derived similarity for the same dishes. Image-level neighborhoods vary substantially across representations, including similarities across culinary-family boundaries. At the dish level, aggregation reveals common visual structure despite local variation, but this structure corresponds less strongly with recipe-derived similarity. These findings show that recognition performance alone cannot characterize relational structure: food similarity depends on representation, modality, and abstraction level rather than being a fixed property of the food collection. Understanding these structures is therefore a prerequisite to combining multiple food information modalities into a coherent notion of similarity.
         
\end{abstract}

\section{Introduction}
\label{sec:intro}

Food is a challenging domain for visual representation learning. Unlike many conventional object categories, its visual appearance is shaped by preparation, ingredients, presentation, regional practice, and cultural context, both within and across dish concepts \cite{bossard2014food,min2023large}. Visual appearance also does not map cleanly onto culinary identity: dishes that look alike may belong to different culinary traditions, while closely related foods may vary substantially in appearance.

Food vision research has nevertheless been dominated by recognition-oriented evaluation. Standard benchmarks ask whether a model can assign the correct dish label to an image, with performance summarized through classification or same-class retrieval accuracy \cite{bossard2014food,min2020isia,min2023large}. Such evaluation establishes whether a representation preserves food identity, but provides limited insight into how it organizes relationships among food concepts. This distinction matters because many food computing applications depend not only on recognizing foods, but on relating them. Retrieval, recommendation, menu exploration, ingredient substitution, and dietary assistance require identifying foods that are meaningfully related to a query. Yet asking whether two foods are similar is underspecified without asking in what respect. Food similarity can arise from appearance, ingredients, preparation, nutrition composition, culinary role, or cultural context, and these relationships need not coincide \cite{salvador2017learning,marin2021recipe1mplus,wang2021cross, min2025multimodal}. The challenge is therefore not simply to learn an embedding in which similar foods are close, but to understand what kind of similarity a representation makes available for comparison.

While a complete account of food similarity will require multiple modalities, we use the visual modality as the primary controlled setting for studying how representations induce similarity structures. We additionally examine recipe text at the dish-concept level, using the same dish concepts, to ask whether a different modality of food information induces a similar relational structure. We study this problem in Indian cuisine, whose regional and culinary diversity produces substantial variation in preparation and presentation, as well as cases where visual appearance and culinary relatedness diverge. This makes Indian cuisine a useful setting for examining how similarity structures vary across representations and modalities, and how visual neighborhoods relate to an independently defined culinary structure.

We therefore ask: \textbf{what notions of food similarity are induced by different representations?} Holding the dataset, query, and retrieval procedure fixed, we compare the neighborhoods produced by seven visual representations and examine them through three contrast cases: cross-family visual similarity, within-family visual dissimilarity, and within-family visual similarity. We then compare the resulting visual similarity structures with recipe-derived similarity over the same dish concepts. Our contributions are as follows:

\begin{enumerate}
    \item \textbf{An expanded Indian food image collection} created by augmenting Khana \cite{prabhu2025khana} with additional Indian food categories, increasing coverage from 80+ classes and 130,000+ images to 300+ classes and 200,000+ images.
    \item \textbf{A retrieval-centric evaluation protocol for visual food similarity} that characterizes representation-induced relationships through image-to-image retrieval, image-to-dish neighborhoods, culinary-family composition, relationship-aware contrast cases, and cross-representation neighborhood agreement.
    \item \textbf{A comparative analysis of seven visual representations} spanning convolutional, transformer-based, and self-supervised architectures, evaluated under a common dataset, similarity function, retrieval procedure, and culinary relationship structure. The analysis shows that different representations can induce substantially different similarity structures over the same collection of foods, as illustrated by the differing nearest-neighbor neighborhoods and similarity scores for a common query in Figure~\ref{fig:biryani_models}.
    \item \textbf{A dish-level cross-modal analysis} that aggregates image instances into dish-to-dish visual structures across representations and compares these structures with recipe-derived similarity over the same dish concepts.
\end{enumerate}


\vspace{-15mm}

\noindent
\begin{minipage}[t][0.55\textheight][c]{0.48\textwidth}
\centering

\fbox{%
\begin{minipage}[t]{0.96\linewidth}
\centering

\includegraphics[
    width=\linewidth,
    height=0.46\textheight,
    keepaspectratio
]{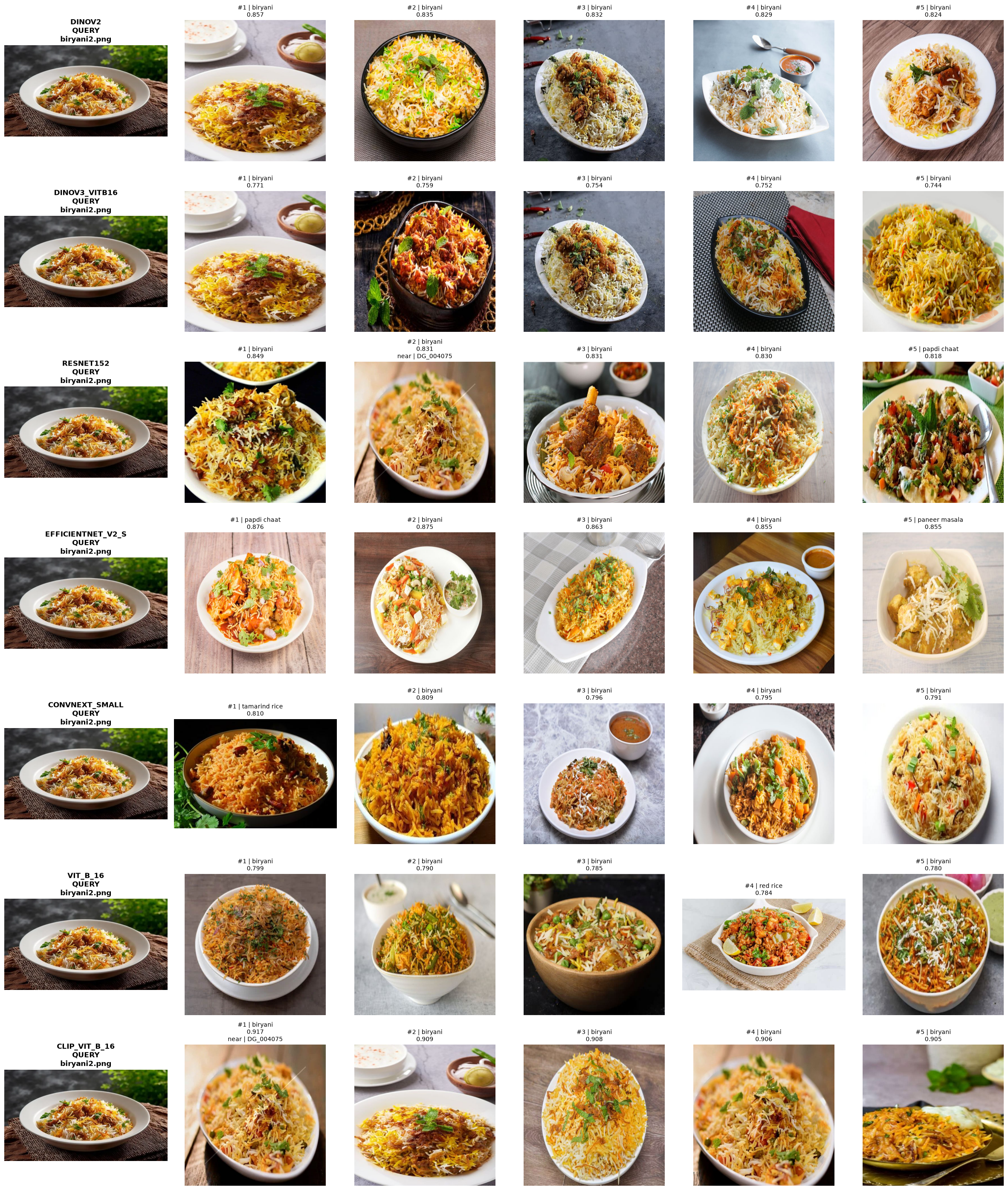}

\end{minipage}
}

\captionof{figure}{
    Nearest neighbors for a \textit{biryani} query across seven
    visual representations.
}
\label{fig:biryani_models}

\end{minipage}
\hfill
\begin{minipage}[t][0.55\textheight][c]{0.48\textwidth}
\centering

\fbox{%
\begin{minipage}[t]{0.96\linewidth}
\centering


\begin{minipage}[t]{0.45\linewidth}
    \centering
    \includegraphics[
        width=0.55\linewidth
    ]{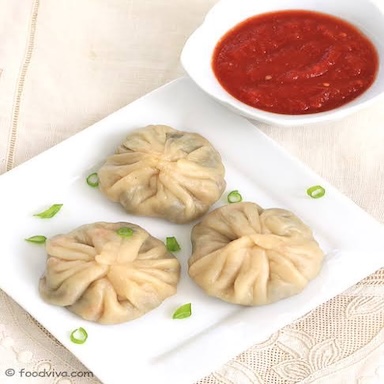}\\[-1pt]
    \scriptsize Steamed \textit{Momo}
\end{minipage}
\hfill
\begin{minipage}[t]{0.45\linewidth}
    \centering
    \includegraphics[
        width=0.55\linewidth
    ]{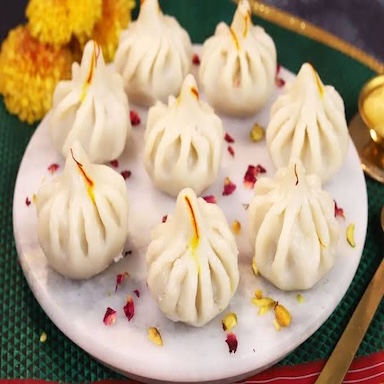}\\[-1pt]
    \scriptsize \textit{Modak}
\end{minipage}

\vspace{1mm}

\scriptsize\textbf{Case A. Cross-family similarity}

\vspace{3mm}


\begin{minipage}[t]{0.45\linewidth}
    \centering
    \includegraphics[
        width=0.55\linewidth
    ]{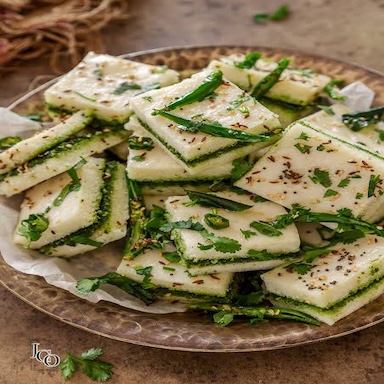}\\[-1pt]
    \scriptsize \textit{Dhokla}
\end{minipage}
\hfill
\begin{minipage}[t]{0.45\linewidth}
    \centering
    \includegraphics[
        width=0.55\linewidth
    ]{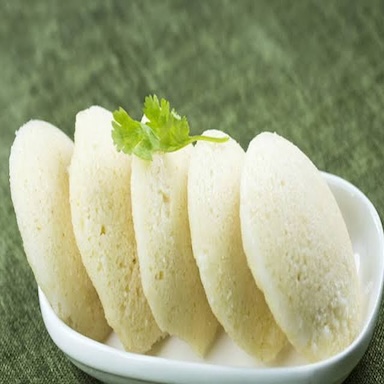}\\[-1pt]
    \scriptsize \textit{Idli}
\end{minipage}

\vspace{1mm}

\scriptsize\textbf{Case B. Within-family dissimilarity}

\vspace{3mm}


\begin{minipage}[t]{0.45\linewidth}
    \centering
    \includegraphics[
        width=0.55\linewidth
    ]{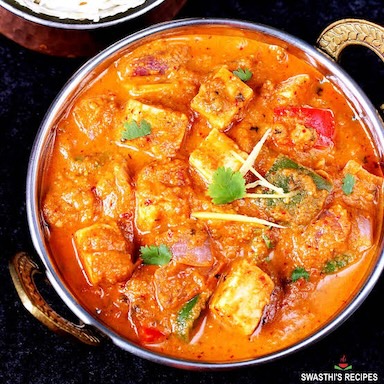}\\[-1pt]
    \scriptsize \textit{Kadhai Paneer}
\end{minipage}
\hfill
\begin{minipage}[t]{0.45\linewidth}
    \centering
    \includegraphics[
        width=0.55\linewidth
    ]{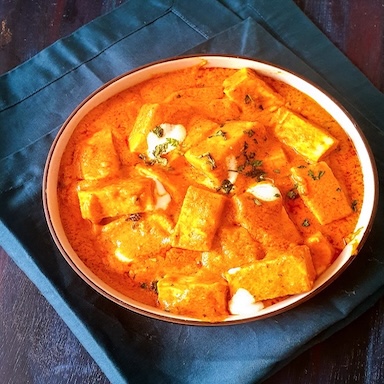}\\[-1pt]
    \scriptsize \textit{Shahi Paneer}
\end{minipage}

\vspace{1mm}

\scriptsize\textbf{Case C. Within-family similarity}

\end{minipage}
}

\captionof{figure}{
Illustrative contrast cases of visual similarity and culinary relatedness.
}
\label{fig:contrast_cases}

\end{minipage}

\vspace{-15mm}

\section{Related Work}
\label{sec:related_work}

Food vision has historically focused on recognition, with increasingly fine-grained datasets spanning diverse cuisines, food traditions, preparation styles, presentation, and imaging conditions \cite{bossard2014food,matsuda2012recognition,kawano2014food,chen2016deep,kaur2019foodx,min2020isia,min2023large,chen2017chinesefoodnet}. Their evaluation, however, is primarily organized around predefined dish identities: representations are assessed according to whether images belonging to the same annotated category are recognized or retrieved together. Such evaluation establishes category discrimination but provides limited information about how different food concepts are positioned relative to one another.

A smaller body of work has addressed food similarity and retrieval more directly. The FoodDist embedding developed for Yum-Me, for example, is intended to reflect visual food similarity and is evaluated through nearest-neighbor retrieval and recommendation \cite{yang2017yum}. Other approaches exploit fine-grained visual cues or alternative sources of supervision to relate food images, while multimodal work has developed image-recipe representations through Recipe1M and Recipe1M+, including methods based on attention, semantic consistency, self-supervision, and cross-domain or multicultural retrieval \cite{salvador2017learning,marin2021recipe1mplus,wang2021cross,zhu2023cross,wang2025mitigating,wang2019learning}. These systems demonstrate that learned representations can support relationships among images, recipes, and other food information, moving beyond recognition toward relating foods to one another. However, their evaluation remains tied to the particular similarity, retrieval, recommendation, or supervision objective used to construct the representation, and retrieval is typically assessed against predefined correspondences such as an associated recipe or image. The resulting neighborhood structure is therefore less often examined in its own right.

More broadly, visual representation learning has diversified from supervised convolutional models to vision transformers, self-supervised representations, and image--text aligned models, including ResNet, EfficientNet, ConvNeXt, ViT, the DINO family, and CLIP \cite{he2016resnet,tan2019efficientnet,liu2022convnet,dosovitskiy2021vit,caron2021dino,oquab2023dinov2,simeoni2025dinov3,radford2021clip}. These models are not interchangeable feature extractors: they differ in architecture, training objectives, supervision, and pretraining data, and can consequently produce different embedding structures. Representation analysis examines neighborhoods, nearest-neighbor structure, clustering, and relationships between learned feature spaces alongside downstream recognition performance \cite{kornblith2019similarity}. For food, where visual appearance can diverge from culinary identity, these differences provide a basis for examining how the same collection is organized under different representations.

Indian cuisine provides a useful setting for this analysis because of its substantial regional and culinary diversity and the complex relationships among dishes, variants, ingredients, and preparation practices \cite{farzanabiryani}. Despite this diversity, Indian food remains comparatively underrepresented in large-scale food-vision benchmarks, with gaps across and within dish categories. Recent work has begun expanding the range of visual food concepts studied through recognition, dietary assessment, detection, and related tasks \cite{agarwal2023indianfoodnet,alla2025foodlens,prabhu2025khana}. However, existing resources provide a relatively coarse relational structure among food concepts, with limited explicit representation of relationships among regional variants, culinary families, and visually similar foods with distinct culinary identities. Khana provides the largest publicly available visual dataset of Indian food, but its primarily dish-level organization leaves these relationships largely unexplored \cite{prabhu2025khana}.

\section{Food Similarity is Representation-Dependent}
\label{sec: food_similarity}

\subsection{Food Lacks a Canonical Similarity Structure}

Similarity is often treated as an intrinsic property of objects. In retrieval systems, this can lead to the assumption that a sufficiently good representation should recover an underlying similarity relation between items. For food, however, no canonical similarity relation is readily defined, a challenge particularly pronounced in Indian cuisine, where culinary identity and visual appearance frequently diverge. This issue became apparent during the construction of FKG.in, a knowledge graph for Indian food \cite{Gupta2024FKG, Gupta2025FKG}. Consider the \textit{samosa}. Although commonly treated as a single dish concept, foods associated with the \textit{samosa} tradition differ substantially in filling, preparation, form, presentation, culinary role, and regional expression while remaining recognizably related. Conversely, foods with different culinary identities may exhibit substantial visual similarity. This raises a more fundamental question than whether two \textit{samosas} look alike: what makes these foods \textit{samosas} in the first place, and in what sense are they similar to one another? Table \ref{tab:samosa_variants} in Appendix \ref{appendix:samosa_variants} illustrates this variation. Shared food identity does not imply uniformity along any single dimension; this variation is not merely deviation from a prototype, but reflects the diversity with which the concept is realized in practice.

This does not make similarity arbitrary. Rather, similarity is relative to the information through which food is represented, rather than being a single property intrinsic to the food collection. Formally, let $d_i$ and $d_j$ denote two food items, understood generically and independently of modality, and let $f(\cdot)$ be a representation function. Their similarity can be computed over their representations: $S(d_i, d_j) = \operatorname{sim}\left(f(d_i), f(d_j)\right)$. Different choices of $f$ can therefore induce different similarity structures over the same food collection. In this work, $f$ is a visual representation learned by an image encoder, so we study visual similarity as a neighborhood structure induced by a representation and a comparison mechanism, distinct from other forms of food relatedness.

\subsection{Visual Similarity, Culinary Relatedness, and Contrast Cases}
\label{subsec:contrast_cases}

\textbf{Visual similarity} arises from information available in an image, such as color, texture, shape, composition, visible ingredients, and presentation, and may therefore place foods that share surface or structural characteristics close together even when their culinary identities differ. \textbf{Culinary relatedness}, in contrast, is defined independently of visual embedding proximity. We operationalize it through culinary-family annotations, where foods may share a dish lineage, preparation tradition, culinary form, or closely related regional variants despite differing in appearance. Family membership is not treated as semantic ground truth, but as an independently defined relational structure against which visual similarity can be examined.

More broadly, food similarity can be grounded in different relationships. Two foods may be \textbf{visually} similar because they share observable appearance, such as \textit{upma} and \textit{halwa}; \textbf{ingredient-related}, such as \textit{rajma} and \textit{chole}; \textbf{procedurally similar}, such as \textit{khichdi} and \textit{pongal}; or \textbf{functionally similar}, such as \textit{naan} and \textit{paratha}. They may also be related through \textbf{cultural or regional culinary traditions}, such as \textit{modak} and \textit{kozhukattai} or through \textbf{shared preference patterns}, where foods are considered similar because they tend to be preferred or selected by overlapping groups of people, independent of their visual or culinary properties. These are distinct relational bases for considering foods similar rather than different names for a single underlying similarity relation.

Our analysis focuses on the relationship between visual similarity and culinary relatedness. Aggregate retrieval metrics such as Precision@K and Recall@K indicate whether a representation preserves predefined food identities, but provide limited insight into how these two relational structures interact. We therefore use three contrast cases as diagnostic probes of representation-induced similarity, focusing on situations where these relations provide different expectations: \textbf{Case A: Cross-family Visual Similarity}, where different dishes from different culinary families are highly similar visually; \textbf{Case B: Within-family Visual Dissimilarity}, where different dishes belonging to the same culinary family exhibit unusually low visual similarity; and \textbf{Case C: Within-family Visual Similarity}, where different dishes within the same culinary family are highly similar visually. The fourth combination, cross-family visual dissimilarity, is not treated as a contrast case because the two relations are aligned rather than in tension. Illustrative examples are shown in Figure \ref{fig:contrast_cases}.

This distinction also matters beyond visual representations. A multimodal representation may incorporate ingredients, recipes, nutrition, or cultural context, but it still induces a particular similarity structure depending on how these sources of information are encoded and combined. Defining a universal food-similarity score would require specifying in advance which dimensions of food relatedness should count as similarity and how they should be combined. Rather than imposing such a universal target, we keep different forms of relatedness conceptually distinct and examine which relationships a given representation makes visible in its induced neighborhoods.

\section{Methodology}
\label{sec:methodology}

We investigate representation-dependent food similarity through a controlled retrieval experiment, holding the food collection, dish labels, culinary-family annotations, similarity function, and retrieval procedure fixed while varying the representation. The analysis proceeds at three levels. First, we compare individual images through \textbf{image-to-image} retrieval. Second, we estimate the similarity between a query image and each candidate dish by aggregating its similarities to the images belonging to that dish, yielding \textbf{image-to-dish} relationships. Third, we construct one visual representation for each dish by aggregating its image embeddings and compare these representations with one another and with recipe-derived representations, yielding \textbf{dish-to-dish} similarity. This section describes the benchmark, embedding extraction, retrieval procedures, and evaluation.

\subsection{Dataset and Benchmark Construction}

The Khana dataset provides 130,000+ images across 80 Indian food dishes \cite{prabhu2025khana}. We augment this collection with 70,000+ additional web-sourced images spanning 330+ identified Indian food categories. Overlapping categories are reconciled and merged at the dish-concept level. During image mining, we use OpenAI’s CLIP model to verify retrieved images against the desired food, dish, or ingredient and filter irrelevant images, while recognizing that CLIP-based filtering may favor visually prototypical images, potentially limiting dataset diversity. The collection is organized at the dish level, retaining multiple images per dish to capture visual variation. Each image is assigned a canonical \textit{dish\_class} representing its food concept. Before embedding extraction, the collection is validated and cleaned: unreadable or corrupted files and non-image files are removed, exact duplicates are removed through image-level hashing, and visually near-identical images are filtered using perceptual similarity procedures. We additionally define the \textit{culinary\_family} relation independently of visual embeddings to capture broader culinary relationships among distinct dish concepts. The final benchmark contains 200,000+ images across more than 300 dish classes and 45 culinary families.

We constructed culinary families as intermediate-level groupings, broader than a dish class but more specific than a generic food category, based on a recognizable common culinary form, preparation tradition, regional or historical lineage, or another culturally established relationship. In assigning families, we considered these dimensions jointly where appropriate, while not treating shared ingredients, isolated cooking techniques, meal role, geographic proximity, or visual similarity alone as sufficient grounds for grouping. The relationship had to be sufficiently specific and unambiguous; otherwise, dishes were kept separate. We retained a family assignment only when its culinary basis met this criterion. Family membership therefore does not imply visual similarity or that member dishes are variants of a named dish. For example, \textit{khichdi} and \textit{pongal} are grouped through their rice-and-lentil preparation traditions; \textit{idli}, \textit{dosa}, \textit{uttapam}, and \textit{paniyaram} through a shared fermented-batter tradition; and \textit{samosa} and \textit{kachori} through a stuffed, deep-fried pastry form. Family names are immaterial to the analysis; the family membership defines the relational structure. These annotations were independently reviewed by two annotators, with disagreements resolved through discussion, and support three levels of comparison: same dish, same culinary family but different dish, and different culinary family. Dish classes without a defensible family relationship remain singleton families and therefore do not contribute to same-family/different-dish comparisons.

\subsection{Visual Representations and Retrieval Structure}

We compare seven pretrained visual representations: ResNet-152 \cite{he2016resnet}, EfficientNet-V2-S \cite{tan2019efficientnet}, ConvNeXt-Small \cite{liu2022convnet}, ViT-B/16 \cite{dosovitskiy2021vit}, DINOv2 \cite{oquab2023dinov2}, DINOv3 \cite{simeoni2025dinov3}, and CLIP ViT-B/16 \cite{radford2021clip}. These span supervised convolutional, transformer-based, self-supervised, and image-text aligned representations; the purpose is to compare substantially different representation spaces rather than establish a model ranking. Each image is processed using the corresponding model preprocessing and mapped to an embedding $z_i^{(m)} = f_m(x_i)$, where $x_i$ denotes the $i$-th image, $f_m$ denotes the embedding function of representation $m$, and $z_i^{(m)}$ denotes the resulting embedding. Embeddings are L2-normalized before similarity computation, with the same image identifiers and ordering maintained across representations. For each representation, images are ranked by cosine similarity to a query image, computed as the dot product: $S_m(x_i,x_j) = \hat{z}^{(m)T}_i \hat{z}^{(m)}_j$.

All eligible images are used as queries. The query image and all images belonging to its identified duplicate or near-duplicate group are excluded from retrieval. Image-level retrieval is performed to a depth of 4,000 candidates and is used directly for same-dish retrieval evaluation. We then derive distinct-dish neighborhoods from these image-level retrievals. The ranked results are collapsed to dish classes by retaining the highest-ranked occurrence of each dish class, so that each dish contributes at most one neighbor. This produces an image-to-dish neighborhood in which the similarity of a candidate dish is determined by its strongest retrieved image. Distinct-dish neighborhoods are evaluated at \(K\in\{5,10,20\}\), with \(K=10\) used as the primary evaluation depth.

We additionally test whether these image-to-dish relationships change with how evidence is aggregated across images within a candidate dish. For each query image and candidate dish, we consider three aggregation rules: the maximum image-level similarity, the mean of the five highest image-level similarities, and the mean similarity across all images in the dish. These alternatives derive dish-level relationships from image-level similarities; they do not construct independent dish-level representations. For these analyses, the top 100 candidate dishes are retained, with \(K=10\) used for evaluation.

\subsection{Dish-Level Visual and Recipe Similarity}

We examine whether the representation dependence observed in the visual modality extends to a different source of food information. At the dish-concept level, we aggregate image embeddings within each dish class using their mean to obtain one visual representation per dish, yielding a dish-to-dish similarity structure over the 307 canonical dish concepts. This aggregation provides a tractable summary of each dish’s visual representation but may obscure within-dish variation. We compare these visual similarity structures across representations and subsequently with the corresponding recipe-derived similarity structure. For the latter, we associate one manually aligned recipe with each dish class, represented using its title, ingredients, and preparation instructions, encoded with \textit{all-mpnet-base-v2}. The analysis is conducted entirely at the dish-concept level, using one recipe embedding per dish; no image-level recipe embeddings or image-to-recipe retrieval are used.

\subsection{Evaluation Metrics}

We evaluate the three levels of similarity separately. For image-to-image retrieval, we report Precision@K, Recall@K, and Mean Reciprocal Rank (MRR), measuring whether retrieved images belong to the query image's predefined dish class. For image-to-dish relationships, we evaluate the composition of the top-\(K\) distinct-dish neighborhoods. We measure the proportion of neighbors belonging to the same culinary family but a different dish, the proportion belonging to different culinary families, and a rank-weighted (RW) family measure. These measures are computed for the primary distinct-dish neighborhoods and for the alternative image-to-dish aggregation rules. Query-level metrics are first averaged within dish class and then macro-averaged across dish classes, so that classes with more images do not receive greater weight in the dataset-level estimates. At the dish-concept level, we compare the \(307\times307\) visual and recipe similarity matrices using pairwise Spearman correlation and top-10 neighborhood Jaccard overlap. We also examine the recipe-induced neighborhoods using the same culinary-family annotations used in the visual analysis.

To examine whether different visual representations organize the same food collection similarly, we compare their distinct-dish retrieval neighborhoods under the same query and evaluation settings. For query $q$, let $N_K^m(q)$ denote the top-$K$ distinct-dish neighborhood induced by representation $m$. Neighborhood overlap between representations $m$ and $n$ is measured as: $O_{m,n}(q) =
\frac{|N_K^m(q) \cap N_K^n(q)|}{K}$. We also compare the ordering of retrieved dish classes using Spearman rank correlation. For each query-model pair, the union of the two top-$K$ dish neighborhoods is considered, with a dish absent from one neighborhood assigned rank $K+1$ for that representation. The primary representation-agreement analysis uses the strongest-image distinct-dish neighborhoods. The same analysis is repeated for the alternative image-to-dish aggregation rules to test whether representation dependence is robust to how image-level evidence is collapsed to dishes.

\subsection{Contrast-Case Operationalization}

The contrast cases in Section~\ref{subsec:contrast_cases} use representation-specific similarity thresholds. For each representation, up to 100,000 valid pairs are sampled per relationship type after excluding duplicate-group pairs. Thresholds are calibrated from the pairwise similarity distributions. \textbf{Case A} comprises different-dish, cross-family pairs with similarity at or above the 99th percentile of the corresponding cross-family distribution. \textbf{Case B} comprises different-dish, same-family pairs with similarity at or below the 10th percentile of the corresponding same-family distribution. \textbf{Case C} comprises different-dish, same-family pairs with similarity at or above the 90th percentile of that distribution. Thresholds are calibrated separately for each representation because similarity distributions may differ across embedding spaces. The resulting cases therefore identify unusually high or low similarities relative to each representation's induced distribution rather than imposing a common absolute cosine-similarity threshold across models.

\section{Results and Discussion}

\subsection{Representation-dependent image retrieval and neighborhoods}

Table~\ref{tab:retrieval_results} shows substantial variation in same-dish retrieval across representations. Precision@10 ranges from 0.258 to 0.523 and MRR from 0.509 to 0.707, while Recall@10 remains low because it is computed over all eligible images of the query dish. Thus, representations differ not only in the images they retrieve, but also in how consistently they place images of the same dish near the query. More importantly, same-dish retrieval does not imply broader culinary relatedness: Family@10 ranges from 0.150 to 0.242, whereas 0.758--0.850 of distinct-dish neighbors belong to different culinary families. Thus, preserving dish identity and organizing broader food relationships are distinct properties of a representation.

\begin{table}[t]
\centering
\caption{Retrieval and neighborhood results at $K=10$ across representations.}
\label{tab:retrieval_results}
\resizebox{\columnwidth}{!}{
\begin{tabular}{l|ccc|ccc|ccc}
\toprule
& \multicolumn{3}{c|}{\textbf{Image-to-Image Retrieval}}
& \multicolumn{3}{c|}{\textbf{Image-to-Dish Neighborhood}}
& \multicolumn{3}{c}{\textbf{Contrast Cases}} \\
\textbf{Model}
& Precision & Recall & MRR
& Family & Cross-family & RW-family
& Case A & Case B & Case C \\
\midrule
ResNet-152 & 0.258 & 0.0078 & 0.509 & 0.150 & 0.850 & 0.166 & 0.766 & 0.000017 & 0.145 \\
EfficientNet-V2-S & 0.277 & 0.0082 & 0.524 & 0.158 & 0.842 & 0.176 & 0.705 & 0.000006 & 0.148 \\
ConvNeXt-Small & 0.305 & 0.0092 & 0.556 & 0.164 & 0.836 & 0.184 & 0.727 & 0.000 & 0.157 \\
ViT-B/16 & 0.282 & 0.0083 & 0.527 & 0.157 & 0.843 & 0.176 & 0.717 & 0.000004 & 0.149 \\
DINOv2 & 0.429 & 0.0136 & 0.643 & 0.215 & 0.785 & 0.241 & 0.740 & 0.000 & 0.204 \\
DINOv3 & 0.523 & 0.0174 & 0.707 & 0.242 & 0.758 & 0.271 & 0.709 & 0.000 & 0.218 \\
CLIP ViT-B/16 & 0.433 & 0.0139 & 0.663 & 0.219 & 0.781 & 0.244 & 0.592 & 0.00019 & 0.185 \\
\bottomrule
\end{tabular}%
}
\end{table}

\subsection{Visual similarity crosses culinary boundaries}

\begin{figure}[t]
    \centering
    \begin{minipage}[t]{0.48\columnwidth}
        \centering
        \setlength{\fboxsep}{2pt}
        \fbox{%
            \includegraphics[
                width=\linewidth,
                height=0.92\linewidth
            ]{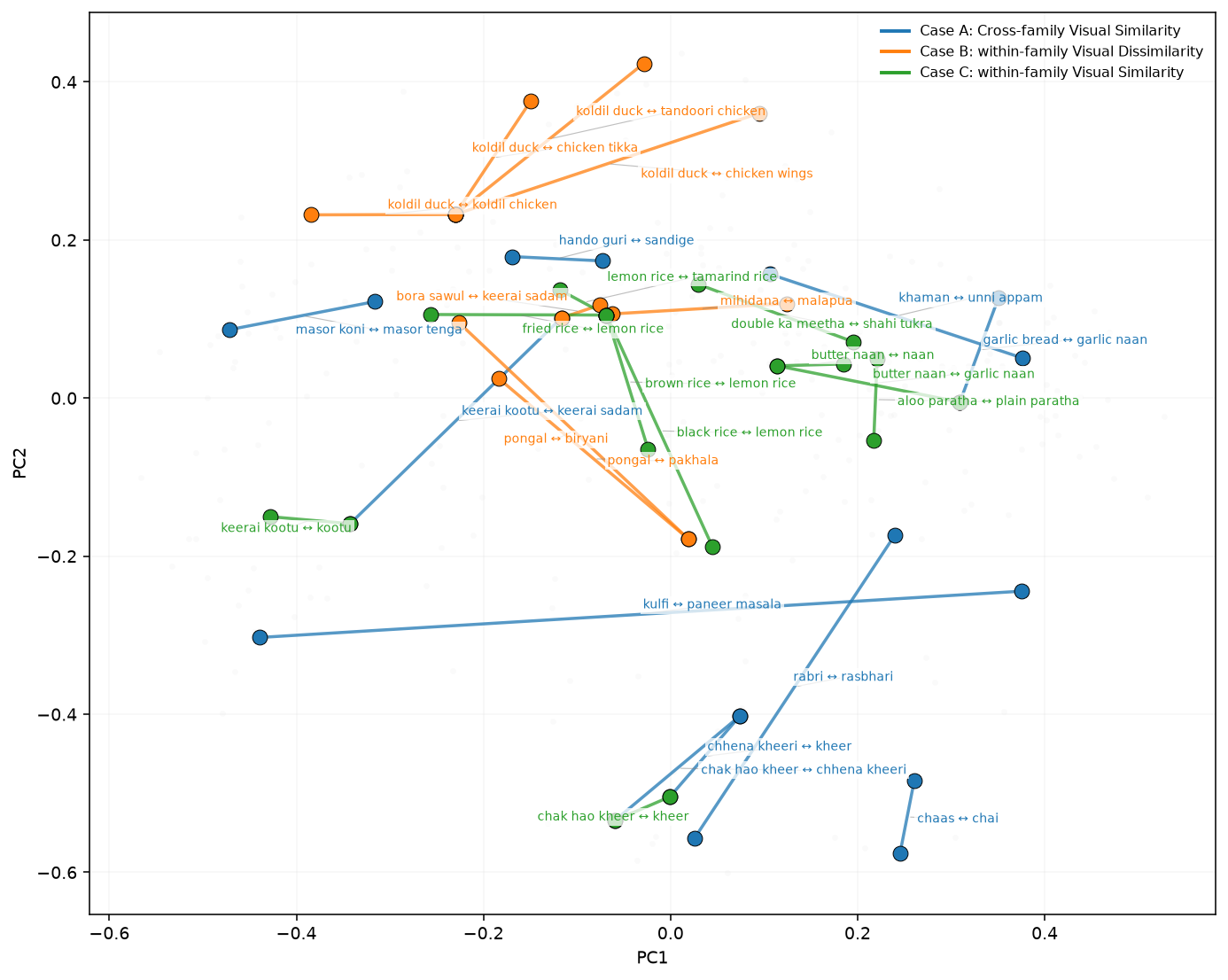}%
        }
        \caption{Visual similarity and culinary relatedness in ViT-B/16. PCA projection with contrast cases highlighted;
        cases are defined in the original embedding space.}
        \label{fig:contrast_case_embedding}
    \end{minipage}
    \hfill
    \begin{minipage}[t]{0.48\columnwidth}
        \centering
        \setlength{\fboxsep}{2pt}
        \fbox{%
            \includegraphics[
                width=\linewidth,
                height=0.92\linewidth
            ]{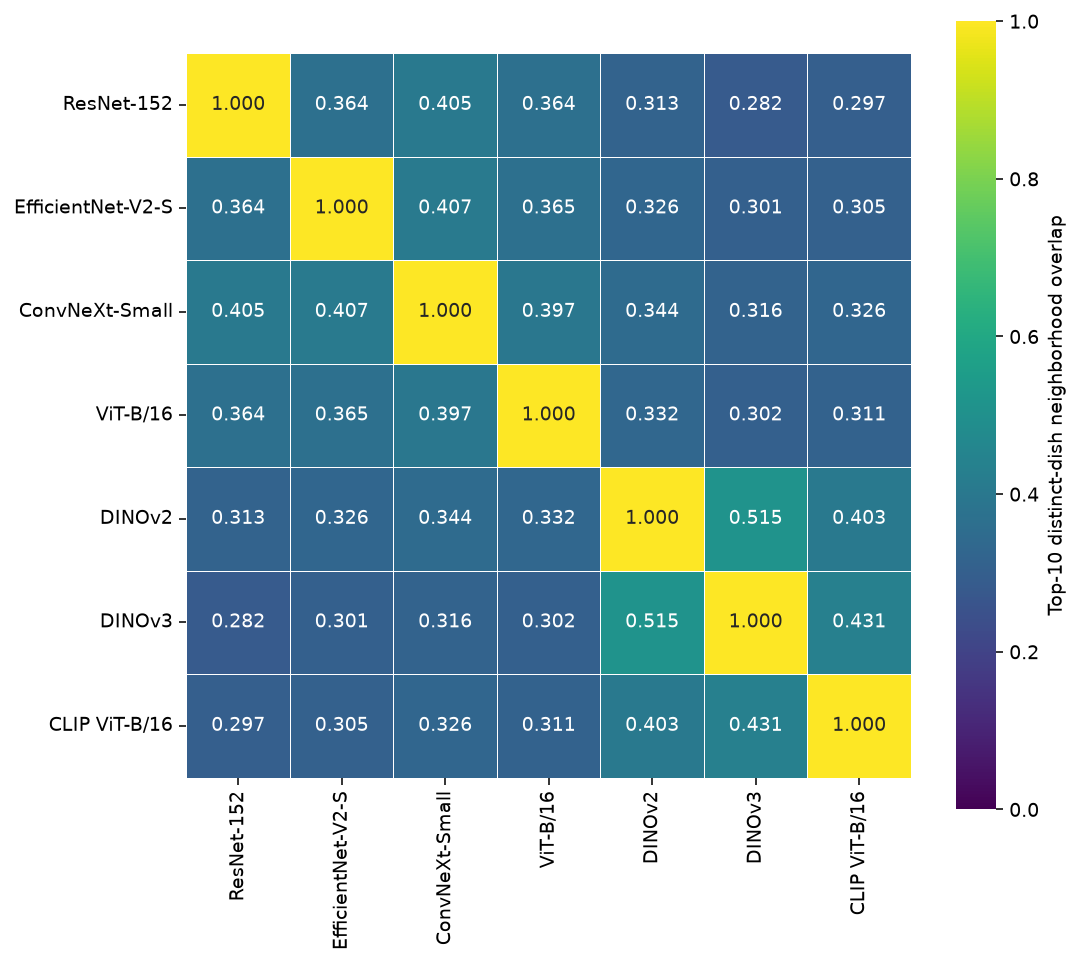}%
        }
        \caption{Pairwise Top-10 neighborhood overlap across
        representations. Values show the mean fraction of shared
        distinct-dish neighbors across query dishes.}
        \label{fig:neighborhood_overlap}
    \end{minipage}
\end{figure}

The contrast cases in Table~\ref{tab:retrieval_results} reveal that visual similarity and culinary-family structure are related but not equivalent. \textbf{Case A}, representing highly similar cross-family foods, occurs at rates of 0.592--0.766, showing that high visual proximity can cross culinary-family boundaries. Conversely, \textbf{Case C}, representing highly similar foods within the same family, occurs at 0.145--0.218, indicating that culinary relatedness can coincide with visual proximity but does not uniquely determine it. \textbf{Case B}, representing unusually dissimilar foods within the same culinary family, is extremely rare across representations and is therefore treated as a diagnostic rather than a central result. Thus, while culinary relatedness does not guarantee visual proximity, members of the same family are also rarely separated by extreme visual dissimilarity. Taken together, these cases suggest that culinary-family structure places some constraints on the extremes of visual organization, while leaving substantial room for visual similarity to cut across family boundaries. Figure~\ref{fig:contrast_case_embedding} illustrates all three relationships in the ViT-B/16 space.

\subsection{The induced neighborhoods differ across representations}

\begin{figure}[t]
    \centering

    \begin{subfigure}[t]{0.48\columnwidth}
        \centering
        \setlength{\fboxsep}{2pt}
        \fbox{%
            \includegraphics[
                width=\linewidth,
                height=0.9\linewidth
            ]{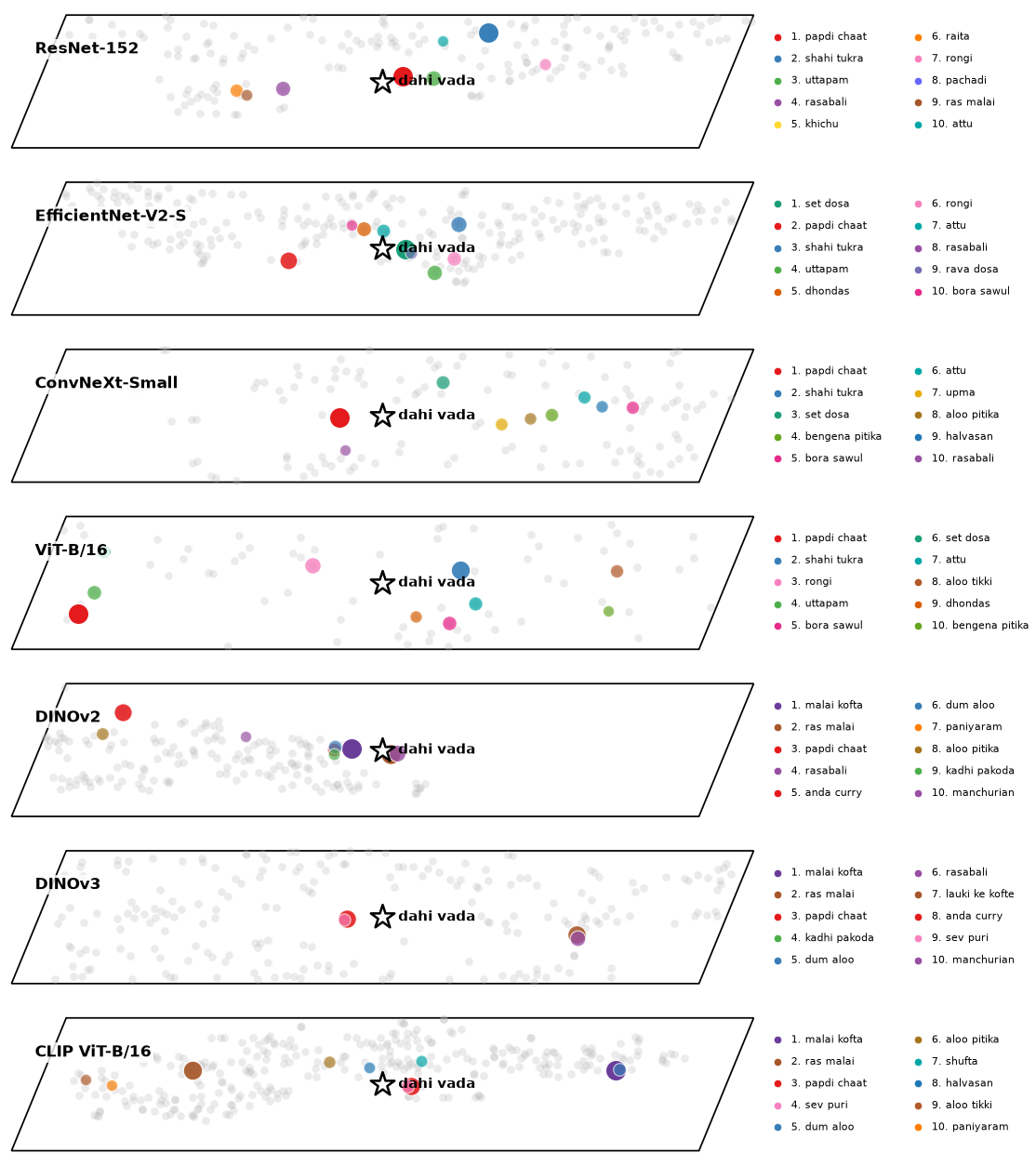}%
        }
        \caption{\textbf{Representation-specific Top-10 neighborhoods of \textit{dahi vada}.} 
        The star marks the query; colored points are retrieved dish concepts,
        with size proportional to image-level similarity. Colors identify dishes,
        numbered lists show retrieval rank, and gray points provide surrounding
        dish-level t-SNE context.}
        \label{fig:dish_neighborhoods}
    \end{subfigure}
    \hfill
    \begin{subfigure}[t]{0.48\columnwidth}
        \centering
        \setlength{\fboxsep}{2pt}
        \fbox{%
            \includegraphics[
                width=\linewidth,
                height=0.9\linewidth
            ]{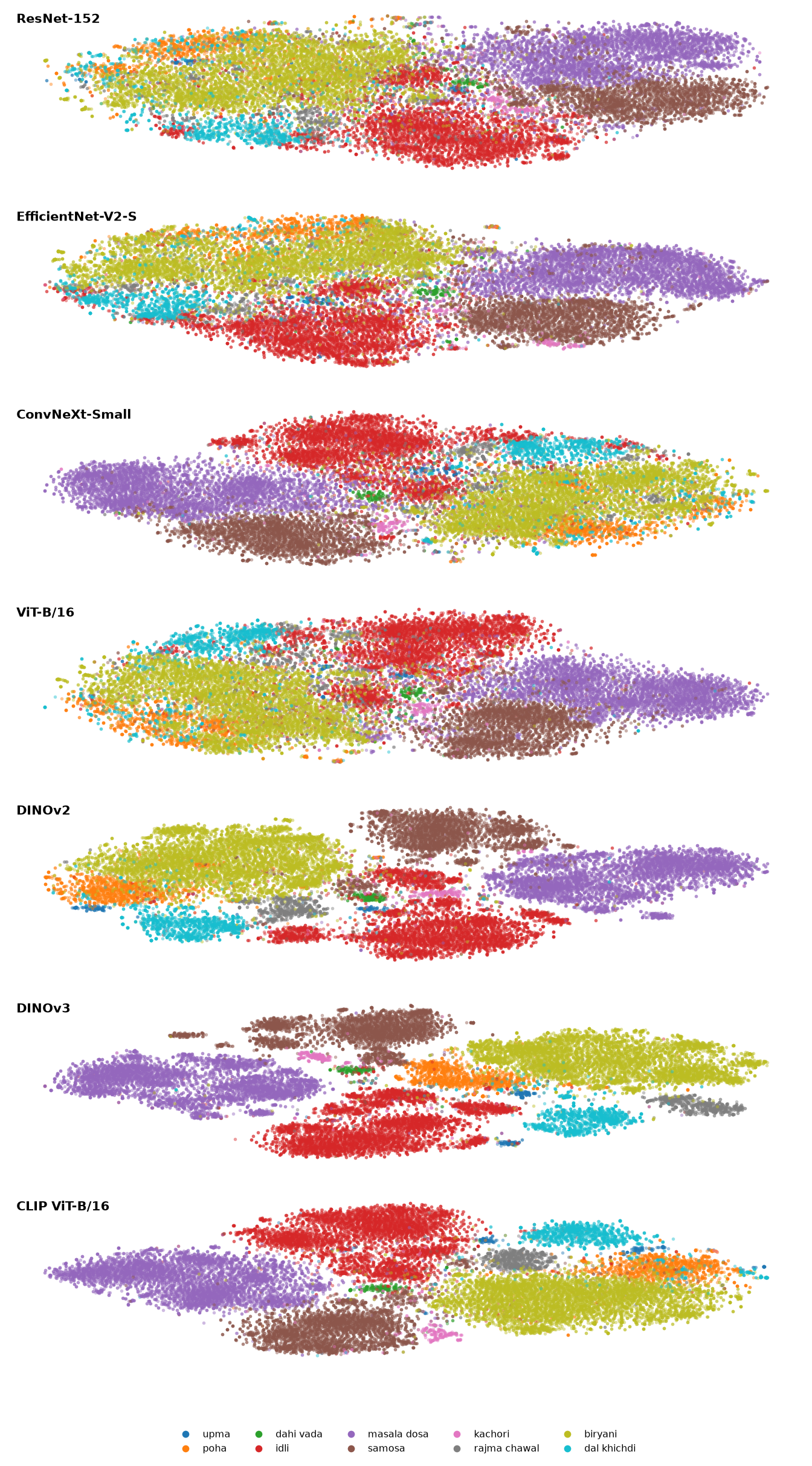}%
        }
        \caption{\textbf{Representation-dependent organization of food images.} t-SNE projections of image instances from ten selected dish concepts reveal differences in cluster compactness, separation, overlap, and fragmentation. Absolute distances and positions are not comparable across independently optimized projections.}
        \label{fig:dish_clusters}
    \end{subfigure}

    \caption{Visualizations of food similarity and organization across seven representations.}
    \label{fig:neighborhoods_and_clusters}

\end{figure}

Figure~\ref{fig:neighborhoods_and_clusters}(a) shows that the Top-10 distinct-dish neighborhood of the same query, \textit{dahi vada}, changes substantially across representations. Figure~\ref{fig:neighborhood_overlap} quantifies this variation: pairwise Top-10 neighborhood overlap ranges from 0.282 to 0.515. Agreement in neighborhood membership is therefore incomplete, and the corresponding rank correlations are generally low, indicating that representations also differ in the ordering of the neighbors they share. Thus, representation dependence operates at two levels: representations can differ in which food concepts they place near a query and in how they order concepts that they retrieve in common. Figure~\ref{fig:neighborhoods_and_clusters}(b) provides a complementary image-level view of this representation-dependent organization. Nevertheless, the representations exhibit partial agreement in their neighborhoods, indicating that they share some visual relationships while differing in the particular relationships they emphasize.

\subsection{Image-to-dish aggregation affects dish-level neighborhoods}

\begin{figure}[t]
    \centering
    \begin{minipage}[t]{0.48\columnwidth}
        \centering
        \setlength{\fboxsep}{2pt}
        \fbox{%
            \includegraphics[
                width=\linewidth,
                height=0.92\linewidth
            ]{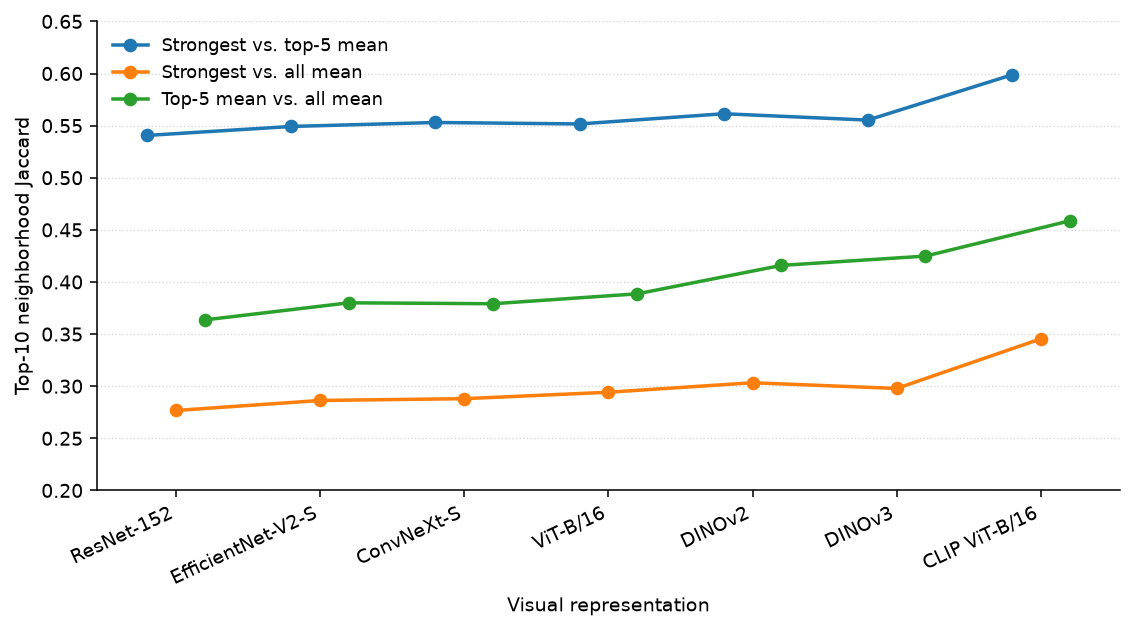}%
        }
        \caption{Image-to-dish neighborhood agreement under alternative aggregation.}
        \label{fig:aggregation_sensitivity}
    \end{minipage}
    \hfill
    \begin{minipage}[t]{0.48\columnwidth}
        \centering
        \setlength{\fboxsep}{2pt}
        \fbox{%
            \includegraphics[
                width=\linewidth,
                height=0.92\linewidth
            ]{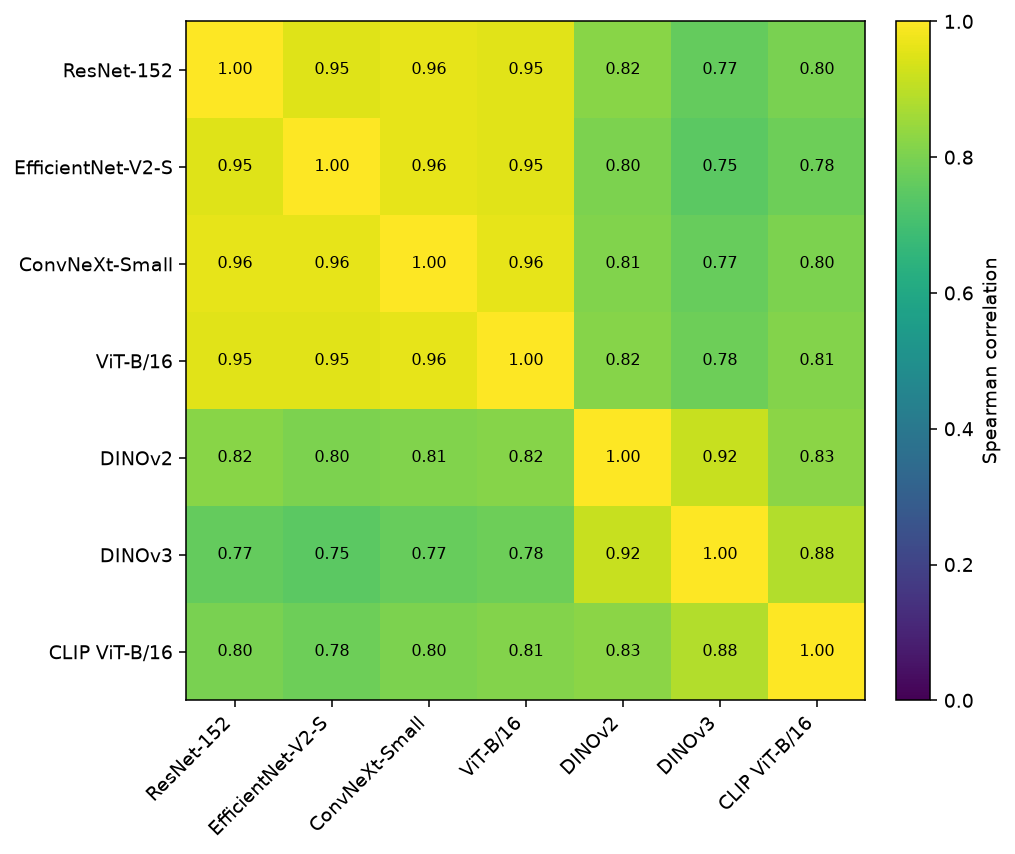}%
        }
        \caption{Cross-representation agreement between dish-level visual structures.}
        \label{fig:spearman_heatmap}
    \end{minipage}
\end{figure}

Figure~\ref{fig:aggregation_sensitivity} shows that dish-level neighborhoods also depend on how image-level evidence is aggregated. Strongest-image and top-five-mean neighborhoods have Top-10 overlap of 0.541--0.599, with approximately 69--73\% of strongest-image relationships recovered by the top-five-mean construction. Thus, a substantial portion of the observed dish relationships is not dependent on a single highly similar image. In contrast, strongest-image versus all-image-mean overlap is only 0.277--0.345, indicating that aggregation affects the resulting neighborhood, particularly when a dish contains substantial visual variation. Taken together, these results show that a limited set of visually representative instances can exert substantial influence on dish-level relationships.

\subsection{Visual and recipe-derived dish similarity}

Aggregating image embeddings within each of the 307 dish classes produces one visual representation per dish and a $307\times307$ similarity structure for each model. Despite representation-dependent neighborhoods at the image level, these dish-level visual structures are strongly correlated across representations, with pairwise Spearman correlations ranging from 0.749 to 0.964 (Figure~\ref{fig:spearman_heatmap}). Thus, representation dependent image-level variation coexists with a substantial common structure at the dish-concept level. The visual structures show substantially weaker correspondence with recipe-derived similarity. As shown in Table~\ref{tab:cross_modal}, visual--recipe Spearman correlations range from 0.223 to 0.326, while Top-10 neighborhood Jaccard overlap ranges from 0.069 to 0.095. This correspondence also varies across visual representations, indicating that recipe-derived structure is related to, but not uniformly reflected in, the visual spaces. Table~\ref{tab:visual_recipe_examples} illustrates dish pairs where visual and recipe similarity, measured by cosine similarity between the corresponding dish-level embeddings, diverge in either direction. Thus, the visual representations share a strong common structure with one another, but only partial correspondence with recipe-derived similarity, which should therefore not be treated as ground-truth culinary similarity.

\begin{figure*}[t]
\centering

\begin{minipage}[t]{0.47\textwidth}
\centering
\captionof{table}{Visual--recipe correspondence.}
\label{tab:cross_modal}
\small
\begin{tabular}{lcc}
\toprule
Model & Spearman $\rho$ & Jaccard@10 \\
\midrule
ResNet-152 & 0.233 & 0.072 \\
EfficientNet-V2-S & 0.251 & 0.069 \\
ConvNeXt-Small & 0.268 & 0.076 \\
ViT-B/16 & 0.260 & 0.078 \\
DINOv2 & 0.223 & 0.079 \\
DINOv3 & 0.281 & 0.089 \\
CLIP ViT-B/16 & 0.326 & 0.095 \\
\bottomrule
\end{tabular}
\end{minipage}
\hfill
\begin{minipage}[t]{0.47\textwidth}
\centering
\captionof{table}{Visual--recipe divergence.}
\label{tab:visual_recipe_examples}
\small
\begin{tabular}{lcc}
\toprule
Dish pair & Visual & Recipe \\
\midrule
\textit{Milk cake} -- \textit{Sukhdi} & 0.917 & 0.349 \\
\textit{Milk cake} -- \textit{Mohanthal} & 0.914 & 0.377 \\
\textit{Ledikeni} -- \textit{Pinaca} & 0.979 & 0.454 \\
\textit{Chakali} -- Kaathi rolls & 0.224 & 0.752 \\
\textit{Adhirasam} -- \textit{Kaathi rolls} & 0.251 & 0.725 \\
\textit{Vada pav} -- \textit{Veg kolhapuri} & 0.329 & 0.790 \\
\textit{Adhirasam} -- \textit{Aloo bonda} & 0.719 & 0.659 \\
\bottomrule
\end{tabular}
\end{minipage}

\end{figure*}

\vspace{\baselineskip}

Taken together, the results show that food similarity is not a single invariant structure recovered by a visual encoder. Representation choice affects image-level neighborhoods, while the way image evidence is aggregated affects dish-level relationships. Similarity in food computing should therefore be understood as a relation induced by a representation and comparison procedure, whose appropriateness depends on the intended retrieval context.

\section{Limitations and Future Work}

This study has several limitations. First, the analysis is conditioned on the image collection and its annotations. Despite cleaning, the collection contains variation in image quality, composition, and source, which can affect absolute retrieval metrics. The results should therefore be interpreted primarily as a controlled comparison of representation-induced structure rather than a benchmark of absolute food-recognition performance. Second, the main analysis is visual. Although the dish-level recipe analysis provides an initial cross-modal comparison, culinary relatedness is broader than either visual or recipe similarity and may also involve other forms of food knowledge. Third, culinary-family assignments necessarily involve abstraction and judgment. Although defined independently of the visual representations and reviewed by two annotators, they capture only selected forms of culinary relatedness. Finally, the recipe analysis uses a single recipe and a fixed text encoder for each of 307 dish concepts, limiting its ability to capture within-dish recipe variation; its results should therefore not be interpreted as ground-truth culinary similarity.

Future work will extend this analysis to richer multimodal representations of Indian food, incorporating images, recipes, ingredients, preparation, nutrition, flavors, and contextual information. Rather than assuming that additional modalities resolve similarity ambiguity, these representations should be evaluated according to the relational structures they induce. Future work could also assess the sensitivity of dish-level similarity to aggregation methods and examine alternative forms of culinary relatedness, rather than treating it as a single dimension. A further direction is controllable or query-conditioned similarity, in which retrieval can explicitly emphasize dimensions such as appearance, ingredients, preparation, nutrition, or culinary tradition. Such a framework would move the question from ``which representation best captures food similarity?'' to ``similar according to which aspects of food, in what ways, and for what purpose?''

\section{Conclusion}

Food similarity is not an invariant property recovered by a model. Across the same food collection and retrieval protocol, changing the visual representation changes both retrieval behavior and the resulting food neighborhoods. These neighborhoods also intersect culinary structure in different ways, showing that visual proximity and culinary relatedness are related but distinct. At the dish level, aggregated visual similarity exhibits a strong common structure across representations, but only partial correspondence with recipe-derived similarity.

For food computing, this means that the challenge is not merely learning better food embeddings or adding more modalities, but understanding what notion of similarity a representation encodes for retrieval. Effective food retrieval depends on which relationships a representation makes accessible through its neighborhoods, not on a single notion of similarity inherent to the food itself. Future food representations should therefore be evaluated not only by whether they preserve food identity, but by the relational structures they induce and whether those structures are appropriate for the intended retrieval context.

\section{Acknowledgments}
This research was supported by the Vachani School of Advanced Computing and the Koita Centre for Digital Health at Ashoka University.

%
%

\bibliographystyle{splncs04}
\bibliography{main}

\begin{thebibliography}{10}
\providecommand{\url}[1]{\texttt{#1}}
\providecommand{\urlprefix}{URL }
\providecommand{\doi}[1]{https://doi.org/#1}

\bibitem{agarwal2023indianfoodnet}
Agarwal, R., Choudhury, T., Ahuja, N.J., Sarkar, T.: Indianfoodnet: Detecting indian food items using deep learning. International Journal of Computational Methods and Experimental Measurements  \textbf{11}(4),  221--232 (2023)

\bibitem{alla2025foodlens}
Alla, J., Samaga, Y., Vaswani, A., Netrapalli, P., Kapania, S., Kumar, P., Hegde, N.: Foodlens: Fine-grained and multi-label classification of indian food images. In: International Conference on Image Processing and Vision Engineering. pp. 262--284. Springer (2025)

\bibitem{bossard2014food}
Bossard, L., Guillaumin, M., Van~Gool, L.: Food-101--mining discriminative components with random forests. In: European conference on computer vision. pp. 446--461. Springer (2014)

\bibitem{caron2021dino}
Caron, M., Touvron, H., Misra, I., J{\'e}gou, H., Mairal, J., Bojanowski, P., Joulin, A.: Emerging properties in self-supervised vision transformers. In: 2021 IEEE/CVF international conference on computer vision (ICCV). pp. 9630--9640. IEEE (2021)

\bibitem{chen2016deep}
Chen, J., Ngo, C.W.: Deep-based ingredient recognition for cooking recipe retrieval. In: Proceedings of the 24th ACM international conference on Multimedia. pp. 32--41 (2016)

\bibitem{chen2017chinesefoodnet}
Chen, X., Zhu, Y., Zhou, H., Diao, L., Wang, D.: Chinesefoodnet: A large-scale image dataset for chinese food recognition. arXiv preprint arXiv:1705.02743  (2017)

\bibitem{dosovitskiy2021vit}
Dosovitskiy, A.: An image is worth 16x16 words: Transformers for image recognition at scale. arXiv preprint arXiv:2010.11929  (2020)

\bibitem{Gupta2024FKG}
Gupta, S.K., Dey, L., Das, P.P., Jain, R.: Building fkg.in: A knowledge graph for indian food. In: Proceedings of the Joint Ontology Workshops (JOWO) -- Episode X: The Tukker Zomer of Ontology, and Satellite Events Co-located with the 14th International Conference on Formal Ontology in Information Systems (FOIS 2024). CEUR Workshop Proceedings, vol.~3882. CEUR-WS.org (2024), \url{https://ceur-ws.org/Vol-3882/ifow-4.pdf}, iFOW: The Integrated Food Ontology Workshop, Enschede, The Netherlands, July 15--19, 2024

\bibitem{Gupta2025FKG}
Gupta, S.K., Dey, L., Das, P.P., Trilok-Kumar, G., Jain, R.: Enhancing fkg.in: Automating indian food composition analysis. In: Pattern Recognition. ICPR 2024 International Workshops and Challenges: Kolkata, India, December 1, 2024, Proceedings, Part IV. pp. 76--90. Berlin, Heidelberg (2025). \doi{10.1007/978-3-031-88217-3_6}

\bibitem{he2016resnet}
He, K., Zhang, X., Ren, S., Sun, J.: Deep residual learning for image recognition. In: Proceedings of the IEEE conference on computer vision and pattern recognition. pp. 770--778 (2016)

\bibitem{kaur2019foodx}
Kaur, P., Sikka, K., Wang, W., Belongie, S., Divakaran, A.: Foodx-251: a dataset for fine-grained food classification. arXiv preprint arXiv:1907.06167  (2019)

\bibitem{kawano2014food}
Kawano, Y., Yanai, K.: Food image recognition with deep convolutional features. In: Proceedings of the 2014 ACM international joint conference on pervasive and ubiquitous computing: Adjunct publication. pp. 589--593 (2014)

\bibitem{kornblith2019similarity}
Kornblith, S., Norouzi, M., Lee, H., Hinton, G.: Similarity of neural network representations revisited. In: International conference on machine learning. pp. 3519--3529. PMLR (2019)

\bibitem{liu2022convnet}
Liu, Z., Mao, H., Wu, C.Y., Feichtenhofer, C., Darrell, T., Xie, S.: A convnet for the 2020s. In: 2022 IEEE/CVF conference on computer vision and pattern recognition (CVPR). pp. 11966--11976. IEEE (2022)

\bibitem{marin2021recipe1mplus}
Marín, J., Biswas, A., Ofli, F., Hynes, N., Salvador, A., Aytar, Y., Weber, I., Torralba, A.: Recipe1m+: A dataset for learning cross-modal embeddings for cooking recipes and food images. IEEE Transactions on Pattern Analysis and Machine Intelligence  \textbf{43}(1),  187--203 (2021). \doi{10.1109/TPAMI.2019.2927476}

\bibitem{matsuda2012recognition}
Matsuda, Y., Hoashi, H., Yanai, K.: Recognition of multiple-food images by detecting candidate regions. In: 2012 IEEE international conference on multimedia and expo. pp. 25--30. IEEE (2012)

\bibitem{min2025multimodal}
Min, W., Hong, X., Liu, Y., Huang, M., Jin, Y., Zhou, P., Xu, L., Wang, Y., Jiang, S., Rui, Y.: Multimodal food learning. ACM Transactions on Multimedia Computing, Communications and Applications  \textbf{21}(7),  1--28 (2025)

\bibitem{min2020isia}
Min, W., Liu, L., Wang, Z., Luo, Z., Wei, X., Wei, X., Jiang, S.: Isia food-500: A dataset for large-scale food recognition via stacked global-local attention network. In: Proceedings of the 28th ACM International Conference on Multimedia. pp. 393--401 (2020)

\bibitem{min2023large}
Min, W., Wang, Z., Liu, Y., Luo, M., Kang, L., Wei, X., Wei, X., Jiang, S.: Large scale visual food recognition. IEEE Transactions on Pattern Analysis and Machine Intelligence  \textbf{45}(8),  9932--9949 (2023)

\bibitem{oquab2023dinov2}
Oquab, M., Darcet, T., Moutakanni, T., Vo, H., Szafraniec, M., Khalidov, V., Fernandez, P., Haziza, D., Massa, F., El-Nouby, A., et~al.: Dinov2: Learning robust visual features without supervision. arXiv preprint arXiv:2304.07193  (2023)

\bibitem{prabhu2025khana}
Prabhu, O.: Khana: A comprehensive indian cuisine dataset. arXiv preprint arXiv:2509.06006  (2025)

\bibitem{radford2021clip}
Radford, A., Kim, J.W., Hallacy, C., Ramesh, A., Goh, G., Agarwal, S., Sastry, G., Askell, A., Mishkin, P., Clark, J., et~al.: Learning transferable visual models from natural language supervision. In: International conference on machine learning. pp. 8748--8763. PMLR (2021)

\bibitem{farzanabiryani}
S, F., Rishi, C.V., Goel, S., Arun, A., Jawahar, C.V.: How does india cook biryani? In: Proceedings of the Sixteen Indian Conference on Computer Vision, Graphics and Image Processing. ICVGIP '25, Association for Computing Machinery, New York, NY, USA (2026). \doi{10.1145/3774521.3774596}

\bibitem{salvador2017learning}
Salvador, A., Hynes, N., Aytar, Y., Marin, J., Ofli, F., Weber, I., Torralba, A.: Learning cross-modal embeddings for cooking recipes and food images. In: Proceedings of the IEEE conference on computer vision and pattern recognition. pp. 3020--3028 (2017)

\bibitem{simeoni2025dinov3}
Sim{\'e}oni, O., Vo, H.V., Seitzer, M., Baldassarre, F., Oquab, M., Jose, C., Khalidov, V., Szafraniec, M., Yi, S., Ramamonjisoa, M., et~al.: Dinov3. arXiv preprint arXiv:2508.10104  (2025)

\bibitem{tan2019efficientnet}
Tan, M., Le, Q.: Efficientnet: Rethinking model scaling for convolutional neural networks. In: International conference on machine learning. pp. 6105--6114. PMLR (2019)

\bibitem{wang2019learning}
Wang, H., Sahoo, D., Liu, C., Lim, E.p., Hoi, S.C.: Learning cross-modal embeddings with adversarial networks for cooking recipes and food images. In: 2019 IEEE/CVF Conference on Computer Vision and Pattern Recognition (CVPR). pp. 11564--11573. IEEE (2019)

\bibitem{wang2021cross}
Wang, H., Sahoo, D., Liu, C., Shu, K., Achananuparp, P., Lim, E.p., Hoi, S.C.: Cross-modal food retrieval: learning a joint embedding of food images and recipes with semantic consistency and attention mechanism. IEEE Transactions on Multimedia  \textbf{24},  2515--2525 (2021)

\bibitem{wang2025mitigating}
Wang, Q., Ngo, C.W., Cao, Y., Lim, E.P.: Mitigating cross-modal representation bias for multicultural image-to-recipe retrieval. In: Proceedings of the 33rd ACM International Conference on Multimedia. pp. 6223--6231 (2025)

\bibitem{yang2017yum}
Yang, L., Hsieh, C.K., Yang, H., Pollak, J.P., Dell, N., Belongie, S., Cole, C., Estrin, D.: Yum-me: a personalized nutrient-based meal recommender system. ACM Transactions on Information Systems (TOIS)  \textbf{36}(1),  1--31 (2017)

\bibitem{zhu2023cross}
Zhu, B., Ngo, C.W., Chen, J., Chan, W.K.: Cross-domain food image-to-recipe retrieval by weighted adversarial learning. arXiv preprint arXiv:2304.07387  (2023)

\end{thebibliography}

\appendix

\section{Samosa Variants}
\label{appendix:samosa_variants}

Table~\ref{tab:samosa_variants} illustrates variation across foods associated with the \textit{samosa} concept.
\begin{table}
\caption{Examples of variation within the \textit{samosa} concept.}
\label{tab:samosa_variants}
\centering
\scriptsize
\setlength{\tabcolsep}{2pt}
\renewcommand{\arraystretch}{0.95}

\begin{tabularx}{\columnwidth}{@{}%
>{\centering\arraybackslash}m{0.12\columnwidth}
>{\raggedright\arraybackslash}m{0.21\columnwidth}
>{\raggedright\arraybackslash}X
>{\raggedright\arraybackslash}m{0.16\columnwidth}
>{\raggedright\arraybackslash}m{0.18\columnwidth}
@{}}
\toprule
Image & Variant & Typical Filling & Preparation & Form \\
\midrule

\includegraphics[height=1.1cm]{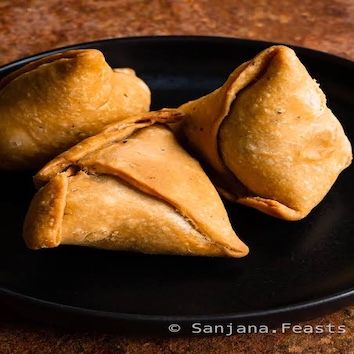}
& Punjabi Samosa
& Potato, peas
& Deep fried
& Large triangular pyramid
\\

\includegraphics[height=1.1cm]{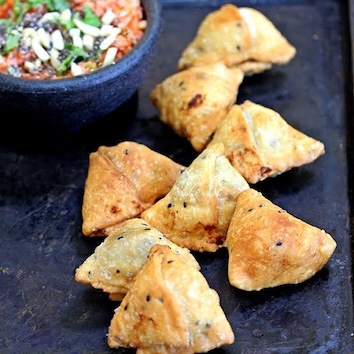}
& Bengali Shingara
& Potato, cauliflower, peanuts
& Deep fried
& Small, flaky triangle
\\

\includegraphics[height=1.1cm]{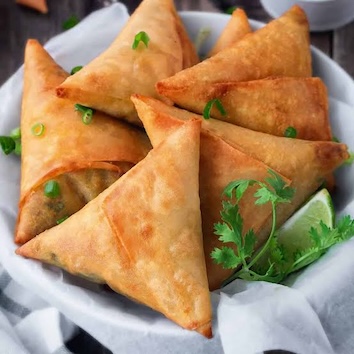}
& Keema Samosa (Mughlai / Bohri)
& Minced meat
& Deep fried
& Triangular; thin wrapper
\\

\includegraphics[height=1.1cm]{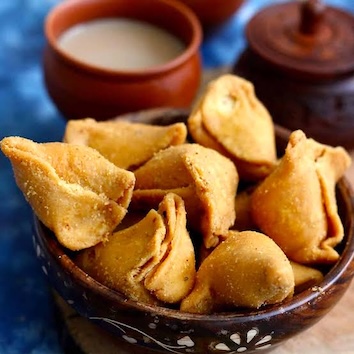}
& Moong Dal Samosa
& Split yellow gram
& Deep fried
& Triangular
\\

\includegraphics[height=1.1cm]{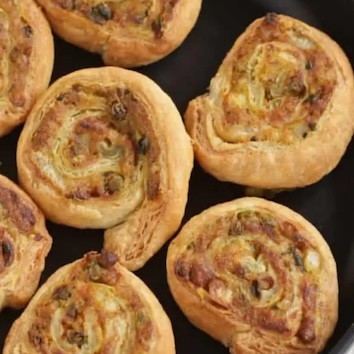}
& Pinwheel Samosa
& Potatoes, green peas
& Deep fried or baked
& Flat spiral wheel
\\

\includegraphics[height=1.1cm]{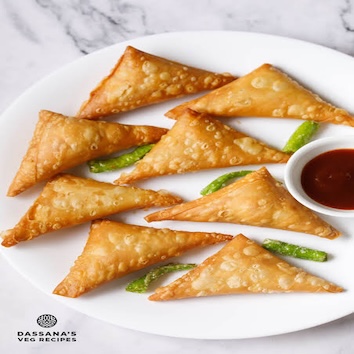}
& Onion Patti Samosa (Irani Samosa)
& Onion, poha
& Deep fried
& Small, thin triangle
\\

\includegraphics[height=1.1cm]{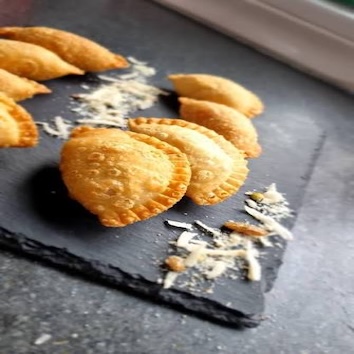}
& Sweet Samosa
& \textit{Khoya}, coconut, nuts
& Deep fried
& Triangle or crescent
\\

\includegraphics[height=1.1cm]{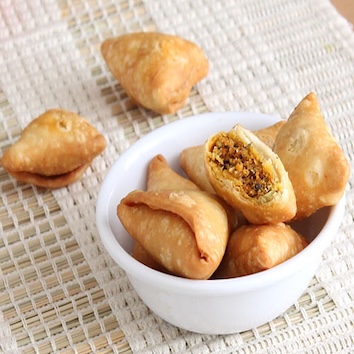}
& Dry Fruit Samosa
& Dry fruits, nuts, sweet filling
& Deep fried
& Small triangle
\\

\includegraphics[height=1.1cm]{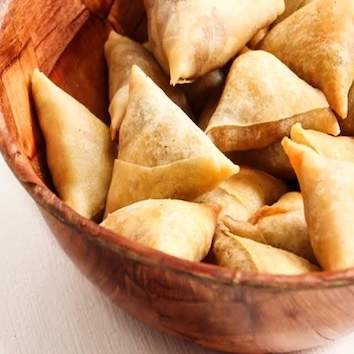}
& Cocktail / Mini Samosa
& Potato, vegetables
& Deep fried
& Miniature triangle
\\

\includegraphics[height=1.1cm]{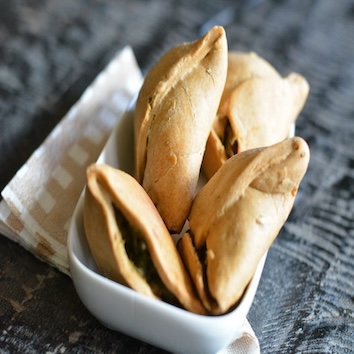}
& Healthy Samosa
& Potato, vegetables, chickpeas, cottage cheese, minced meat
& Baked or air-fried
& Triangular
\\

\includegraphics[height=1.1cm]{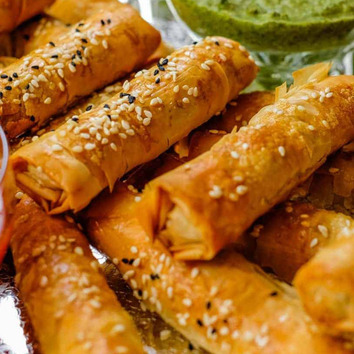}
& Cigar Samosa
& Potato, vegetables, meat
& Deep-fried
& Long, thin cylindrical roll 
\\

\includegraphics[height=1.1cm]{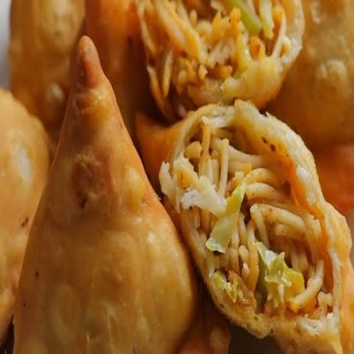}
& Chinese Noodle Samosa
& Noodles, vegetables
& Deep fried
& Triangular
\\

\includegraphics[height=1.1cm]{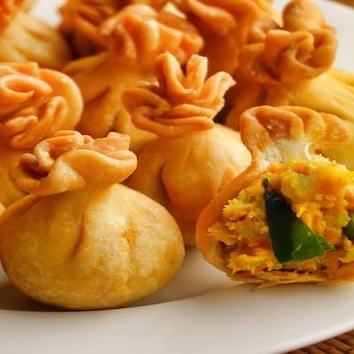}
& Potli Samosa
& Potato, vegetables
& Deep fried
& Pouch-shaped
\\

\includegraphics[height=1.1cm]{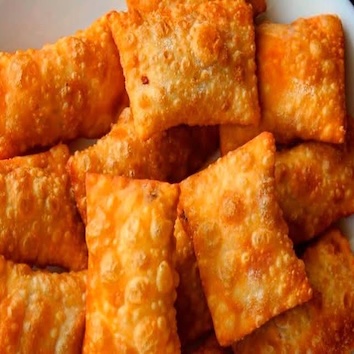}
& Lukhmi Samosa
& Minced meat
& Deep fried
& Flat square / rectangle
\\

\includegraphics[height=1.1cm]{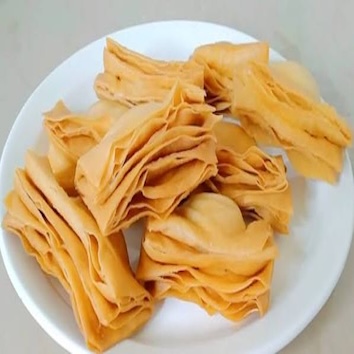}
& Warqi Samosa
& Meat, egg, vegetables
& Deep fried
& Layered, flaky triangle
\\

\includegraphics[height=1.1cm]{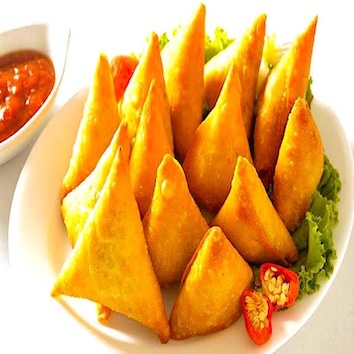}
& Burmese Samusa
& Potato, onion, peas
& Deep fried
& Flat triangle
\\

\includegraphics[height=1.1cm]{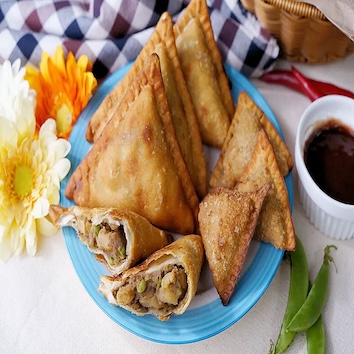}
& Goan Chamuça
& Meat, vegetables
& Deep fried
& Triangular
\\

\includegraphics[height=1.1cm]{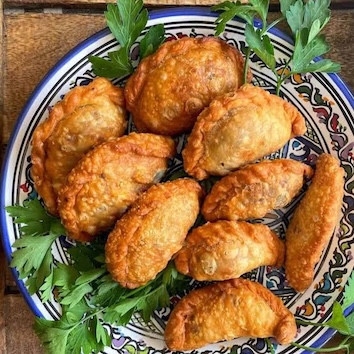}
& Middle Eastern Sambousek
& Meat, cheese, chickpeas, vegetables
& Deep fried or baked
& Crescent, triangular, or tubular
\\

\bottomrule
\end{tabularx}
\end{table}

\end{document}